\def\cvprPaperID{****}
\def\confName{CVPR}
\def\confYear{****}

\def\authorBlock{
    Yufan Ren$^1$\thanks{Equal contribution} \qquad
    Fangjinhua Wang$^2$\footnotemark[1] \qquad
    Tong Zhang$^1$ \qquad
    Marc Pollefeys$^2$ \qquad
    Sabine Süsstrunk$^1$ \\
    $^1$IVRL IC EPFL \qquad $^2$Department of Computer Science, ETH Zurich \\
}

\newif\ifreview 
\newif\ifarxiv \newcommand{\arxiv}{\arxivtrue}
\newif\ifcamera 
\newif\ifrebuttal 

\arxiv

\pdfoutput=3
\documentclass[10pt,twocolumn,letterpaper]{article}
\ifreview \usepackage[review]{cvpr} \fi
\ifarxiv \usepackage[pagenumbers]{cvpr} \fi
\ifrebuttal \usepackage[rebuttal]{cvpr} \fi
\ifcamera \usepackage{cvpr} \fi

\usepackage{graphicx}
\usepackage{amsmath}
\usepackage{amssymb}
\usepackage{booktabs}

\usepackage[pagebackref,breaklinks,colorlinks]{hyperref}
\newcommand\customparagraph[1]{\vspace{0.4em}\noindent\textbf{#1.}}

\usepackage{comment}
\usepackage{widetable}
\usepackage{multirow}
\usepackage{color}

\usepackage{times}
\usepackage{microtype}
\usepackage{epsfig}
\usepackage[table,xcdraw]{xcolor}
\usepackage{caption}
\usepackage{float}
\usepackage{placeins}
\usepackage{color, colortbl}
\usepackage{stfloats}
\usepackage{enumitem}
\usepackage{tabularx}
\usepackage{xstring}
\usepackage{multirow}
\usepackage{xspace}
\usepackage{url}
\usepackage{subcaption}
\usepackage{comment}
\usepackage{xcolor}
\usepackage[hang,flushmargin]{footmisc}
\usepackage{widetable}
\usepackage{makecell}

\ifcamera \usepackage[accsupp]{axessibility} \fi

\ifarxiv  \fi

\newcommand{\R}[1]{{%
    \textbf{%
        \ifstrequal{#1}{1}{\textcolor{red}{R#1}}{%
        \ifstrequal{#1}{2}{\textcolor{blue}{R#1}}{%
        \ifstrequal{#1}{3}{\textcolor{magenta}{R#1}}{%
        \ifstrequal{#1}{4}{\textcolor{teal}{R#1}}{%
                           \textcolor{cyan}{R#1}%
        }}}}%
    }%
}}

\usepackage{xr-hyper}

\makeatletter
\newcommand*{\addFileDependency}[1]{
  \typeout{(#1)}
  \@addtofilelist{#1}
  \IfFileExists{#1}{}{\typeout{No file #1.}}
}

\makeatother

\usepackage[pagebackref,breaklinks,colorlinks]{hyperref}
\usepackage[capitalize]{cleveref}
\crefname{section}{Sec.}{Secs.}
\crefname{table}{Table}{Tables}
\crefname{figure}{Fig.}{Figs.}

\begin{document}

\title{VolRecon: Volume Rendering of Signed Ray Distance Functions \\ for Generalizable Multi-View Reconstruction}
\author{\authorBlock}
\maketitle

\begin{abstract}

The success of the Neural Radiance Fields (NeRF) in novel view synthesis has inspired researchers to propose neural implicit scene reconstruction. 
However, most existing neural implicit reconstruction methods optimize per-scene parameters and therefore lack generalizability to new scenes. 
We introduce VolRecon, a novel generalizable implicit reconstruction method with Signed Ray Distance Function (SRDF). 
To reconstruct the scene with fine details and little noise, VolRecon combines projection features aggregated from multi-view features, and volume features interpolated from a coarse global feature volume. 
Using a ray transformer, we compute SRDF values of sampled points on a ray and then render color and depth. 
On DTU dataset, VolRecon outperforms
SparseNeuS by about 30\% in sparse view reconstruction and
achieves comparable accuracy as MVSNet in full view reconstruction.
Furthermore, our approach exhibits good generalization performance on the large-scale ETH3D benchmark. Code is available at \url{https://github.com/IVRL/VolRecon/}.

\end{abstract}
\section{Introduction}
\label{sec:intro}

The ability to reconstruct 3D geometries from images or videos is crucial in various applications in robotics~\cite{weiss2011monocular,tancik2022block,hane2015obstacle} and augmented/virtual reality \cite{oswald2014generalized,middelberg2014scalable}.
Multi-view stereo (MVS)~\cite{schonberger2016pixelwise,yao2018mvsnet,yao_2019_rmvsnet,gu2020cascade,wang2021patchmatchnet,galliani_2015_gipuma} is a commonly used technique for this task. 
A typical MVS pipeline involves multiple steps, \ie, multi-view depth estimation, filtering, and fusion~\cite{galliani_2015_gipuma,curless1996volumetric}. 

\begin{figure}[!t]
\setlength{\belowcaptionskip}{-0.375cm}

    \centering
    \includegraphics[width=0.5\textwidth,page=1]{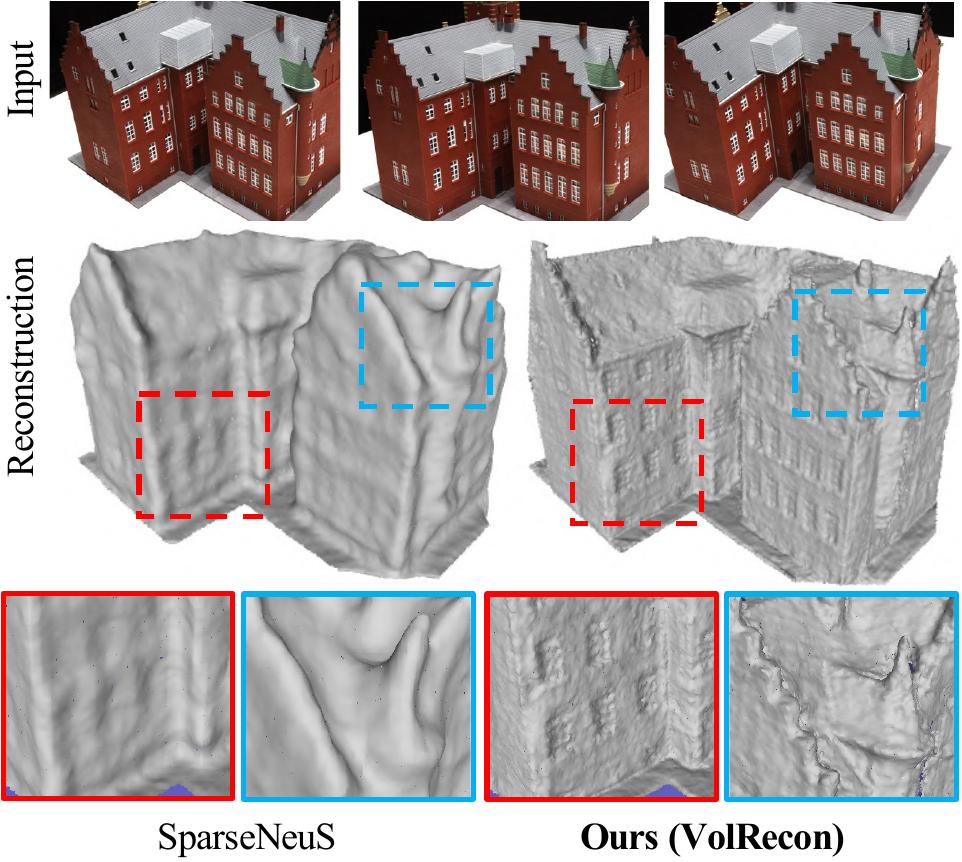}
    
    \caption{Generalizable implicit reconstructions from three views (top). The state-of-the-art method SparseNeuS~\cite{long2022sparseneus} produces over-smoothed surfaces (left), while \textbf{our (VolRecon)} reconstructs finer details (right). Best viewed on a screen when zoomed in.} 

\label{fig:teaser}
\end{figure}

Recently, there has been a growing interest in neural implicit representations for various 3D tasks, such as shape modeling~\cite{mescheder2019occupancy, park2019deepsdf}, surface reconstruction~\cite{yariv2021volume,wang2021neus}, and novel view synthesis~\cite{mildenhall2021nerf}. 
NeRF~\cite{mildenhall2021nerf}, a seminal work in this area, employs Multi-Layer Perceptrons (MLP) to model a radiance field, producing volume density and radiance estimates for a given position and viewing direction. 
While NeRF's scene representation and volume rendering approach has proven effective for tasks such as novel view synthesis, it cannot generate accurate surface reconstruction due to difficulties in finding a universal density threshold for surface extraction~\cite{yariv2021volume}. 
To address this, researchers have proposed neural implicit reconstruction using the Signed Distance Function (SDF) for geometry representation and modeling the volume density function~\cite{wang2021neus,yariv2021volume}. 
However, utilizing SDF with only color supervision leads to unsatisfactory reconstruction quality compared to MVS methods~\cite{yao2018mvsnet,gu2020cascade} due to a lack of geometry supervision and potential radiance-geometry ambiguities~\cite{zhang2020nerf++,wei2021nerfingmvs}. 
As a result, subsequent works have sought to improve reconstruction quality by incorporating additional priors, such as sparse Struction-from-Motion (SfM) point clouds~\cite{fu2022geoneus}, dense MVS point clouds~\cite{zhang2022critical}, normals~\cite{wang2022neuris,yu2022monosdf}, and depth maps~\cite{yu2022monosdf}.

Many neural implicit reconstruction methods are restricted to optimizing one model for a particular scene and cannot be applied to new, unseen scenes,~\ie, across-scene generalization.
However, the ability to generalize learned priors to new scenes is valuable in challenging scenarios such as reconstruction with sparse views~\cite{yu2021pixelnerf,chen2021mvsnerf,wang2021ibrnet}. 
In order to achieve across-scene generalization in neural implicit reconstruction, it is insufficient to simply input the spatial coordinate of a point as NeRF. Instead, we need to incorporate information about the scene, such as the points' projection features on the corresponding images~\cite{yu2021pixelnerf,chen2021mvsnerf,wang2021ibrnet}. 
SparseNeuS~\cite{long2022sparseneus} recently achieved across-scene generalization in implicit reconstruction with global feature volumes~\cite{chen2021mvsnerf}. 
Despite achieving promising results, SparseNeuS is limited by the resolution of the feature volume due to
the memory constraints~\cite{kostrikov2014probabilistic,murez2020atlas}, leading to over-smoothing surfaces even with a higher resolution feature volume, Fig.~\ref{fig:teaser}.

In this paper, we propose VolRecon, a novel framework for generalizable neural implicit reconstruction using the Signed Ray Distance Function (SRDF). 
Unlike SDF, which defines the distance to the nearest surface along any directions, SRDF~\cite{zins2022multi} defines the distance to the nearest surface along a given ray. 
We utilize a projection-based approach to gather local information about surface location. We first project each point on the ray into the feature map of each source view to interpolate multi-view features. Then, we aggregate the multi-view features to \textit{projection features} using a view transformer. 
However, when faced with challenging situations such as occlusions and textureless surfaces, determining the surface location along the ray with only local information is difficult. 
To address this, we construct a coarse global feature volume that encodes global shape priors like SparseNeuS~\cite{murez2020atlas,long2022sparseneus}. 
We use the interpolated features from the global feature volume, \ie, \textit{volume features}, and \textit{projection features} of all the sampled points along the ray to compute their SRDF values, with a ray transformer. 
Similar to NeuS~\cite{wang2021neus}, we model the density function with SRDF and then estimate the image and depth map with volume rendering.

Extensive experiments on DTU~\cite{aanaes2016_dtu} and ETH3D~\cite{2017eth3d} verify the effectiveness and generalization ability of our method. 
On DTU, our method outperforms the state-of-the-art method SparseNeuS \cite{long2022sparseneus} by 30\% in sparse view reconstruction and 22\% in full view reconstruction. %
Furthermore, our method performs better than the MVS baseline COLMAP~\cite{schonberger2016pixelwise}.
Compared with MVSNet~\cite{yao2018mvsnet}, a seminal learning-based MVS method, our method performs better in the depth evaluation and has comparable accuracy in full-view reconstruction.
On the ETH3D benchmark~\cite{2017eth3d}, we show that our method has a good generalization ability to large-scale scenes. 

In summary, our contributions are as follows: %
\begin{itemize}
    \item We propose VolRecon, a new pipeline for generalizable implicit reconstruction that produce detailed surfaces.
    \item Our novel framework comprises a view transformer to aggregate multi-view features and a ray transformer to compute SRDF values of all the points along a ray.
    \item We introduce a combination of local projection features and global volume features, which enables the reconstruction of surfaces with fine details and high quality. 
\end{itemize}

\begin{figure*}[!ht]
    \centering
    \includegraphics[width=1.02\textwidth, page=2]{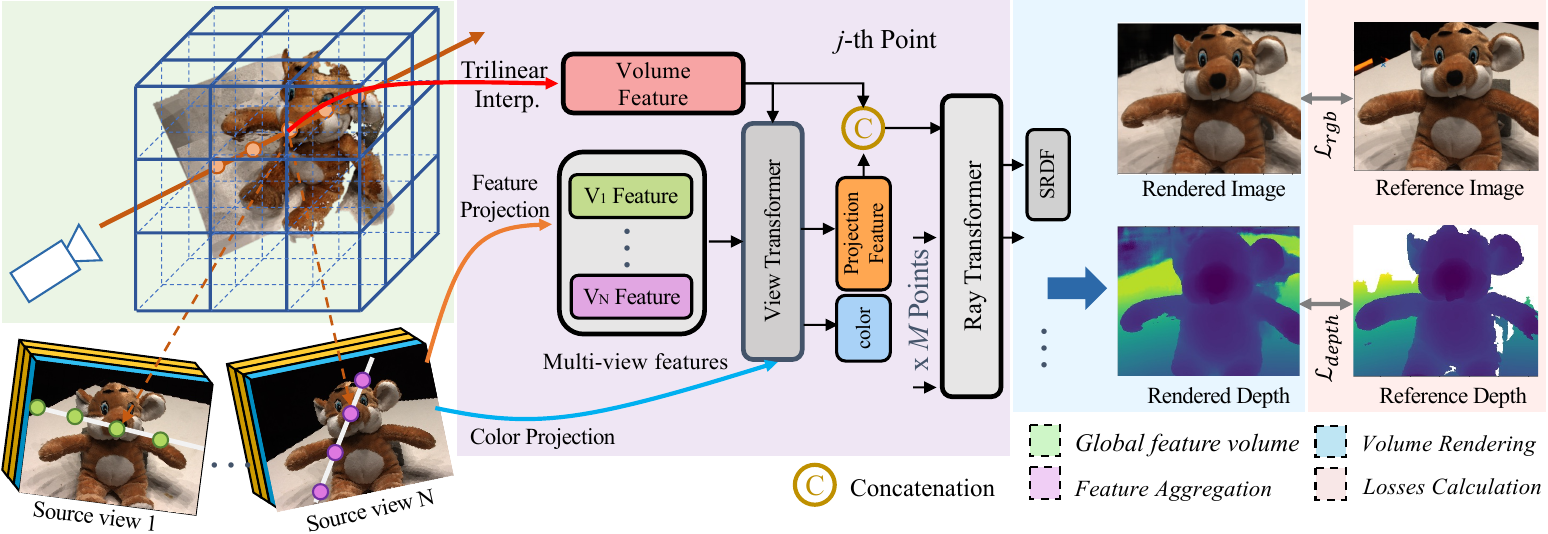}

    \caption{
    \textbf{Structure of VolRecon}. For a set of $N$ source views, we first extract the image features. Then we build a global feature volume to obtain global shape priors. Next, given a ray in the target viewpoint, we project each sampled point on the ray into the source views, aggregate its multi-view features using a view transformer to get the projection feature and blended color of each point. After that, we apply the ray transformer to concatenated features of all the $M$ points along the ray to predict their SRDF values. Finally, we volume render the color and depth. Best viewed on a screen when zoomed in. 
    }

\label{fig:pipeline}
\end{figure*}
\section{Related Work}
\label{sec:related}

\customparagraph{Neural Implicit Reconstruction}
Traditional volumetric reconstructions~\cite{curless1996volumetric,niessner2013real,izadi2011kinectfusion} use implicit signed distance fields to produce high-quality reconstructions. %
Recent works use networks to model shapes as continuous decision boundaries, \ie, occupancy functions~\cite{mescheder2019occupancy,peng2020convolutional} or SDF~\cite{park2019deepsdf}. 
In NeRF~\cite{mildenhall2021nerf}, the authors further show that combining neural implicit functions,~\eg, Multi-Layer Perceptron (MLP), and volume rendering can achieve photo-realism in novel view synthesis~\cite{barron2021mip,mildenhall2021nerf,fridovich2022plenoxels,muller2022instant,chan2021pi}. 
Since NeRF~\cite{mildenhall2021nerf}, which originally targets a per-scene optimization problem, several additional methods~\cite{yu2021pixelnerf,chen2021mvsnerf,wang2021ibrnet} are proposed to perform generalizable novel view synthesis for unseen scenes. 
For example, IBRNet~\cite{wang2021ibrnet} projects sampled points along the ray into multiple source views. It aggregates multi-view features into density features and uses a ray transformer, which inputs the density features for all points along the ray to predict the density for each point. 
For multi-view reconstruction, IDR~\cite{yariv2020multiview} reconstructs surfaces by representing
the geometry as the zero-level set of an MLP, requiring accurate object masks. 
To avoid using masks, VolSDF~\cite{yariv2021volume} and NeuS~\cite{wang2021neus} incorporate SDF in neural volume rendering, using it to modify the density function. 
Additional geometric priors~\cite{fu2022geoneus,zhang2022critical,wang2022neuris,yu2022monosdf} were proposed to improve the reconstruction quality. 
Nevertheless, these methods usually require a lengthy optimization for each scene and cannot generalize to unseen scenes. 

Recently, SparseNeuS~\cite{long2022sparseneus} attempts to solve across-scene generalization for surface reconstruction. 
Similar to~\cite{murez2020atlas,sun2021neuralrecon,chen2021mvsnerf}, 
SparseNeuS constructs fixed-resolution feature volumes to aggregate image features from multi-view images. 
An MLP takes the coordinates and corresponding interpolated features from the feature volumes to predict the SDF values as input. 
SparseNeuS needs high-resolution volumes, \ie, $192^3$, but still outputs over-smoothed surfaces.
In contrast, we additionally use the \textit{projection feature} that contains local features and use a ray transformer to aggregate features of sampled points along a ray.
In this way, our VolRecon model captures both local and global information to achieve finer detail and less noise than SparseNeuS, Fig.~\ref{fig:teaser}. 

\customparagraph{Multi-view Stereo}
Based on scene representations, traditional MVS methods fall into three main categories: 
volumetric~\cite{kutulakos2000theory, seitz1999photorealistic, kostrikov2014probabilistic}, point cloud-based~\cite{lhuillier2005quasi, furukawa2010}, and depth map-based~\cite{galliani_2015_gipuma, schonberger2016pixelwise, xu_2019_acmm}. 
Depth map-based methods are more flexible since they decouple the problem into depth map estimation and fusion~\cite{galliani_2015_gipuma,schonberger2016pixelwise}. 
Therefore, most recent learning-based MVS methods~\cite{yao2018mvsnet,yao_2019_rmvsnet,gu2020cascade,zhang2020vismvsnet,wang2021patchmatchnet,wang2022itermvs} perform multi-view depth estimation and then fuse them to a point cloud, which achieves impressive performance on various benchmarks~\cite{aanaes2016_dtu, knapitsch2017tanks,2017eth3d}. 
Note that while much progress has been made in neural implicit reconstruction, the reconstruction performance~\cite{yariv2021volume,yariv2020multiview,wang2021neus,long2022sparseneus} is still not on par with the state-of-the-art MVS baselines. 
Yet, our method performs better than COLMAP in few view reconstruction and achieves comparable accuracy as MVSNet~\cite{yao2018mvsnet} in full view reconstruction.

\section{Method}
\label{sec:method}

In this section, we discuss the structure of VolRecon, illustrated in Fig.~\ref{fig:pipeline}. 
The pipeline consists of predicting the Signed Ray Distance Function (SRDF) (Sec.~\ref{sec::ray_oriented_srdf}), volume rendering of the SRDF to predict color and depth (Sec.~\ref{sec::volume_rendering}), and loss functions (Sec.~\ref{sec::loss}). 

\subsection{SRDF Prediction} \label{sec::ray_oriented_srdf}

\customparagraph{Signed Ray Distance Function}
Let set $\Omega \in \mathbb{R}^3$ denotes the space and $\mathcal{M}=\partial \Omega$ its boundary surface.
The Signed Distance Function $d_{\Omega}(\mathbf{p})$ defines the shortest distance of a point $\mathbf{p}\in \mathbb{R}^3$ to the surface $\mathcal{M}$. Its sign denotes whether $\mathbf{p}$ is outside (positive) or inside (negative) of the surface,

\begin{equation}
\mathbf{1}_{\Omega}(\mathbf{p})= \begin{cases}1 & \text { if } \mathbf{p} \in \Omega \\ 0 & \text { if } \mathbf{p} \notin \Omega\end{cases},
\end{equation}

\begin{equation}
d_{\Omega}(\mathbf{p})=(-1)^{\mathbf{1}_{\Omega}(\mathbf{p})} \min _{\mathbf{p^*} \in \mathcal{M}} \|\mathbf{p}-\mathbf{p^*}\|_2,
\end{equation}
where $||\cdot||_2$ is the $L_2$-norm and $\mathbf{p^*}$ are points on the surface. Differently, SRDF~\cite{zins2022multi,curless1996volumetric} defines the shortest distance to surface $\mathcal{M}$ along a  ray direction $\mathbf{v}$ ($||\mathbf{v}||_2=1$),

\begin{equation}
\tilde d_{\Omega}(\mathbf{p,v})=(-1)^{\mathbf{1}_{\Omega}(\mathbf{p})} \min _{\mathbf{p^*} \in \mathcal{M},\frac{\mathbf{p^*}-\mathbf{p}}{||\mathbf{p^*}-\mathbf{p}||_2}=\mathbf{v}}\|\mathbf{p}-\mathbf{p^*}\|_2.   
\end{equation}

Theoretically, given a point $\mathbf{p}$, its SDF $d_{\Omega}(\mathbf{p})$ equals to the SRDF $\tilde d_{\Omega}(\mathbf{p,v})$ with the minimum absolute value in any direction $\mathbf{v}$:

\begin{equation}
    d_{\Omega}(\mathbf{p})=(-1)^{\mathbf{1}_{\Omega}({\mathbf{p}})} \min_{\mathbf{v}}(|\tilde d_{\Omega}(\mathbf{p}, \mathbf{v})|).
\end{equation}

Similar to SDF volume rendering~\cite{wang2021neus, yariv2021volume}, we incorporate SRDF in volume rendering to estimate the depth map from the given viewpoints, which can be fused into mesh~\cite{curless1996volumetric} or dense point clouds~\cite{schonberger2016pixelwise}.

\customparagraph{Feature Extraction} \label{sec::feature_extraction}
Given the source image set $\mathbb{I} = \{ \mathbf{I}_1, \cdots, \mathbf{I}_N  \}$, where $\mathbf{I} \in \mathbb{D}^{H \times W \times 3}$, $\mathbb{D} \subset [0,1]$, and $H,W$ are the image height and width, respectively.
We use a Feature Pyramid Network~\cite{lin2017fpn} to extract feature maps $\{\mathbf{F}_i\}_{i=1}^N\in \mathbb{R}^{\frac H4\times \frac W4\times C}$. 

\customparagraph{Global Feature Volume} \label{sec::global_feature_volume}
We construct a global feature volume $\mathbf{F}_v$ similar to~\cite{murez2020atlas, sun2021neuralrecon} to get global information. 
Specifically, we first divide the bounding volume of the scene into $K^3$ voxels. 
The center point of each voxel is projected onto the feature map of each source view to obtain the features. 
This is done using bilinear interpolation, where the mean and variance of $N$ features are computed and concatenated as the voxel features. 
We then use 3D U-Net\cite{ronneberger2015u} to regularize and aggregates the information. 
For each point $\mathbf{p}$, we denote the interpolated feature from $\mathbf{F}_v$ as \textit{volume feature}, $\mathbf{f}_v$. 
Please refer to supplementary for more details. 

\customparagraph{View Transformer} 
Given a pixel in the reference view, we denote the $M$ points on the ray emitted from this pixel as $\{\mathbf{p}(t) = \mathbf{o} + t\mathbf{v}, t \geq 0\}$. 
By projecting each point $\mathbf{p}$ onto the feature map of each source view, we extract colors $\{\mathbf{c}_i\}_{i=1}^{N}$ and features $\{\mathbf{f}_i\}_{i=1}^{N}$ using bilinear interpolation. 
We apply a \textit{view transformer} to aggregate the multi-view features $\{\mathbf{f}_i\}_{i=1}^{N}$ into one feature, which we denote as the \textit{projection feature}. 
Structurally, we use a self-attention transformer~\cite{vaswani2017attention} with linear attention~\cite{katharopoulos2020transformers}.
Following previous work~\cite{dosovitskiy2020image}, we add a learnable aggregation token, denoted as $\mathbf{f}_0$, to obtain the projection feature. 
Since no order of source views is assumed, we do not use positional encoding in the view transformer. 
The projection feature $\mathbf{f}_p$ and updated multi-view features $\{\mathbf{f'}_i\}_{i=1}^{N}$ are computed as, 

\begin{equation}
    \mathbf{f}_p, \{\mathbf{f'}_i\}_{i=1}^{N} = \\ \text{ViewTrans}(\mathbf{f}_0, \{\mathbf{f}_i\}_{i=1}^{N}).
\end{equation}

Visibility is important in multi-view aggregation~\cite{schonberger2016pixelwise,wang2021patchmatchnet} due to the existence of occlusions. 
Therefore, mean and variance aggregation~\cite{wang2021ibrnet,yao2018mvsnet} may not be robust enough since all views are accounted equally. 
Using a learnable transformer enables the model to reason about the consistency for aggregation across multiple views. 

\customparagraph{Ray Transformer} 
Similar to SDF, SRDF is not locally defined and its value depends on the closest surface along the ray. 
To provide such non-local information of other points along the ray, we additionally design a \textit{ray transformer} based on linear attention~\cite{katharopoulos2020transformers}. 
We first concatenate the projection feature and corresponding volume feature into a combined feature to add global shape prior.
After ordering the points in a sequence from near to far, the ray transformer applies positional encoding~\cite{wang2021ibrnet} and self-attention on the combined feature to predict attended features $\{ \mathbf{\tilde{f}}_j\}_{j=1}^{M}$,
\begin{equation}
    \{\mathbf{\tilde{f}}_j\}_{j=1}^{M} = \\ \text{RayTrans}(\{\text{cat}(\mathbf{f}_v,\mathbf{f}_p, \gamma)\}_{j=1}^{M}),
\end{equation}
where $\text{cat}(\cdot)$ denotes concatenation and $\gamma$ positional encoding. 
Finally, we use an MLP to decode the attended feature to SRDF for each point on the ray. 

\subsection{Volume Rendering of SRDF} \label{sec::volume_rendering}

\customparagraph{Color Blending} 
For a point $\mathbf{p}$ at viewing direction $\mathbf{v}$, we blend colors of $N$ source views,
$\{\mathbf{c}_i\}_{i=1}^{N}$,
similar to~\cite{wang2021ibrnet, wang2022generalizable}. 
We compute the blending weight using the updated multi-view features $\{\mathbf{f'}_i\}_{i=1}^{N}$ from the view transformer. 
Similar to~\cite{wang2021ibrnet, wang2022generalizable}, we concatenate $\{\mathbf{f'}_i\}_{i=1}^{N}$ with the difference between $\mathbf{v}$ and the viewing direction in the $i$-th source view, $\mathbf{v}_i$. 
Then we pass the concatenated features through an MLP and use \textit{Softmax} to get the blending weights $\{\eta_i\}_{i=1}^N$. %
The final radiance at point $\mathbf{p}$ and viewing direction $\mathbf{v}$ is the weighted sum of $\{\mathbf{c}_i\}_{i=1}^{N}$, 
\begin{equation}
    \hat{\mathbf{c}} = \sum_{i=1}^{N} \eta_i \cdot \mathbf{c}_i.
\end{equation}

\customparagraph{Volume rendering} 
Several works~ \cite{yariv2021volume, wang2021neus} propose to include SDF in volume rendering for implicit reconstruction with the supervision of the pixel reconstruction loss.
We adopt the method of NeuS~\cite{wang2021neus} to volume render SRDF, as briefly introduced below. We provide a comparison between rendering SDF and SRDF in supplementary. 

Specifically, the color is accumulated along the ray 
\begin{equation} \label{eq::sdfrender}
    \hat{\mathbf{C}}=\sum_{j=1}^M T_j \alpha_j \hat{\mathbf{c}}_j,
\end{equation}
where $T_j=\prod_{k=1}^{j-1}\left(1-\alpha_k\right)$ is the discrete \textit{accumulative transmittance}, and $\alpha_j$ are discrete opacity values defined by
\begin{equation}
    \alpha_j = 1- \exp (-\int_{t_j}^{t_{j+1}} \rho(t) dt),
\end{equation}
where opaque density $\rho(t)$ is similar to the original definition in NeuS~\cite{wang2021neus}. The difference is that we replace the original SDF with SRDF in $\rho(t)$. For more theoretical details, please refer to~\cite{wang2021neus}. 

Similar to volume rendering of colors, we can derive the rendered depth as 

\begin{equation} \label{eq::depthrender}
    \hat{\mathbf{D}}=\sum_{j=1}^M T_j \alpha_j t_j.
\end{equation}

\subsection{Loss Function} \label{sec::loss}
We define the loss function as

\begin{equation}
    \mathcal{L}=\mathcal{L}_{\text {color }}+\alpha \mathcal{L}_{\text {depth}}. %
\end{equation}

The color loss $\mathcal{L}_{\text{color}}$ is defined as
\begin{equation}
    \mathcal{L}_{\text {color}} = \frac{1}{S} \sum_{s=1}^S \left\|\hat{\mathbf{C}}_s-\mathbf{C}_s\right\|_2,
\end{equation}

where $S$ is the number of pixels and $\mathbf{C}_s$ is the ground truth color. 

The depth loss $\mathcal{L}_{\text{depth}}$ is defined as
\begin{equation}
    \mathcal{L}_{\text {depth}} = \frac{1}{S_1} \sum_{s=1}^{S_1} |\hat{\mathbf{D}}_s - \mathbf{D}_s|,
\end{equation}
where $S_1$ is the number of pixels with valid depth and $\mathbf{D}_s$ is the ground truth depth. 
In our experiments, we choose $\alpha=1.0$.

\section{Experiments}

\subsection{Experimental Settings}

\begin{figure*}[t]
    \centering
    \includegraphics[width=1.0\textwidth, page=3]{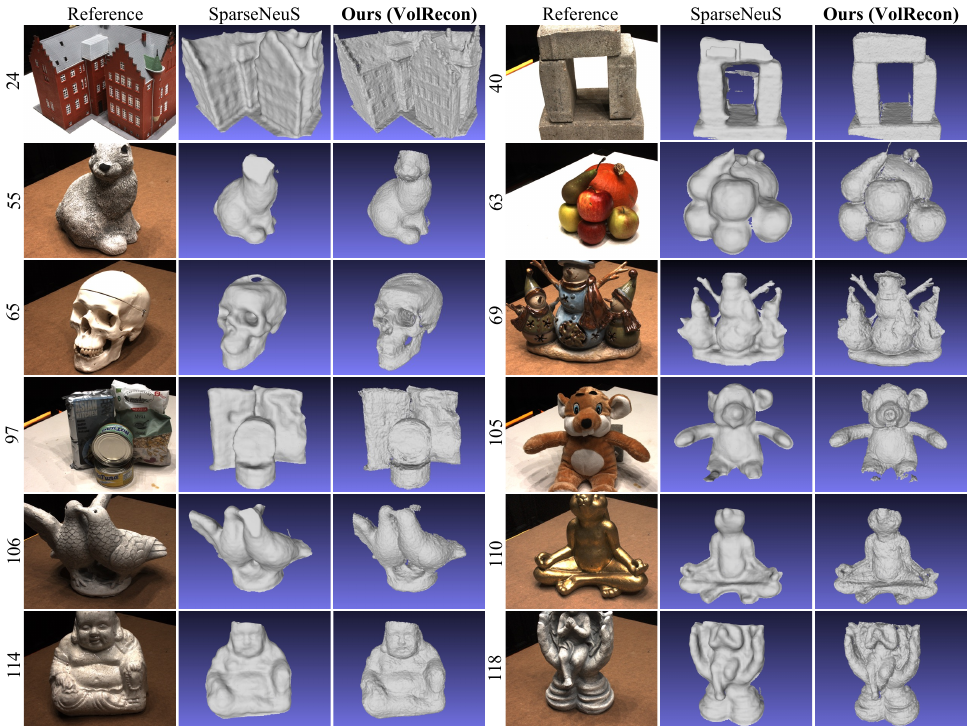}
    \caption{\textbf{Sparse view} ($N=3$) reconstruction on testing scenes in the DTU benchmark~\cite{aanaes2016_dtu}. While SparseNeuS~\cite{long2022sparseneus} produces over-smoothed surfaces, our method (VolRecon) reconstructs with finer details, \eg, scene 24 windows and scene 63 fruit stalks, and sharper boundaries, \eg, scene 97 cans touching part and scene 118 sculpture base, than SparseNeuS~\cite{long2022sparseneus}. Best viewed on a screen when zoomed in. 
    }
\label{fig:sparseview_mesh}
\end{figure*}

\customparagraph{Datasets} 
Following existing works~\cite{wang2021neus,yariv2021volume,fu2022geoneus,long2022sparseneus}, we use the DTU dataset~\cite{aanaes2016_dtu} for training. 
The DTU dataset~\cite{aanaes2016_dtu} is an indoor multi-view stereo dataset with ground truth point clouds of 124 different scenes and 7 different lighting conditions. 
During experiments, we use the same 15 scenes as SparseNeuS for testing and use the remaining scenes for training. 
We use the depth maps rendered from the mesh~\cite{yao2018mvsnet} as depth map ground truth.
Besides DTU, we also use the ETH3D dataset~\cite{2017eth3d} to test the generalization ability of our method. 
ETH3D~\cite{2017eth3d} is a challenging MVS benchmark consisting of high-resolution images of real-world large-scale scenes with strong viewpoint variations.

\begin{table*}[ht]

\resizebox{\textwidth}{!}{%
\renewcommand{\arraystretch}{1.1}
\begin{tabular}{lcccccccccccccccc} \\ \hline
Scan         & \multicolumn{1}{l}{Mean$ \downarrow$} & 24   & 37   & 40   & 55   & 63   & 65   & 69   & 83   & 97   & 105  & 106  & 110  & 114  & 118  & 122  \\ \hline
COLMAP~\cite{schonberger2016pixelwise} & 1.52 & \textbf{0.90} & 2.89 & \underline{1.63} & \underline{1.08} & 2.18 & 1.94 & 1.61 & \underline{1.30} & 2.34 & 1.28 & \underline{1.10} & 1.42 & 0.76 & \underline{1.17} & \textbf{1.14} \\ 
MVSNet~\cite{yao2018mvsnet} & \textbf{1.22} & \underline{1.05} & \textbf{2.52} & 1.71 & \textbf{1.04} & \underline{1.45} & \textbf{1.52} & \textbf{0.88} & \textbf{1.29} & \textbf{1.38} & \textbf{1.05} & \textbf{0.91} & \textbf{0.66} & \textbf{0.61} & \textbf{1.08} & \underline{1.16} \\
\hline
IDR \cite{yariv2020multiview} & 3.39 & 4.01 & 6.40 & 3.52 & 1.91 & 3.96 & 2.36 & 4.85 & 1.62 & 6.37 & 5.97 & 1.23 & 4.73 & 0.91 & 1.72 & 1.26 \\
VolSDF \cite{yariv2021volume} & 3.41 & 4.03 & 4.21 & 6.12 & 0.91 & 8.24 & 1.73 & 2.74 & 1.82 & 5.14 & 3.09 & 2.08 & 4.81 & 0.60 & 3.51 & 2.18 \\
UNISURF \cite{oechsle2021unisurf} & 4.39 & 5.08 & 7.18 & 3.96 & 5.30 & 4.61 & 2.24 & 3.94 & 3.14 & 5.63 & 3.40 & 5.09 & 6.38 & 2.98 & 4.05 & 2.81 \\
NeuS \cite{wang2021neus} & 4.00 & 4.57 & 4.49 & 3.97 & 4.32 & 4.63 & 1.95 & 4.68 & 3.83 & 4.15 & 2.50 & 1.52 & 6.47 & 1.26 & 5.57 & 6.11 \\ \hline

PixelNeRF \cite{yu2021pixelnerf}   & 6.18                     & 5.13 & 8.07 & 5.85 & 4.40  & 7.11 & 4.64 & 5.68 & 6.76 & 9.05 & 6.11 & 3.95 & 5.92 & 6.26 & 6.89 & 6.93 \\
IBRNet \cite{wang2021ibrnet}      & 2.32                     & 2.29 & 3.70  & 2.66 & 1.83 & 3.02 & 2.83 & 1.77 & 2.28 & 2.73 & 1.96 & 1.87 & 2.13 & 1.58 & 2.05 & 2.09 \\
MVSNeRF \cite{chen2021mvsnerf}     & 2.09                     & 1.96 & 3.27 & 2.54 & 1.93 & 2.57 & 2.71 & 1.82 & 1.72 & 2.29 & 1.75 & 1.72 & 1.47 & 1.29 & 2.09 & 2.26 \\ \hline

SparseNeuS \cite{long2022sparseneus}   & 1.96                     & 2.17 & 3.29 & 2.74 & 1.67 & 2.69 & 2.42 & 1.58 & 1.86 & 1.94 & 1.35 & 1.50  & 1.45 & 0.98 & 1.86 & 1.87 \\
\textbf{Ours (VolRecon)} & \underline{1.38}   & 1.20	& \underline{2.59} &	\textbf{1.56}	& \underline{1.08} &	\textbf{1.43} &	\underline{1.92} &	\underline{1.11} & 1.48 & \underline{1.42} & \textbf{1.05} & 1.19 & \underline{1.38} & \underline{0.74} & 1.23 & 1.27  \\ \hline 
\end{tabular}
}
\caption{Quantitative results of \textbf{sparse view} reconstruction on 15 testing scenes of DTU dataset~\cite{aanaes2016_dtu}. We report Chamfer distance (lower is better).  
Methods are separated into four categories (from top to bottom): (1) multi-view stereo (MVS) baselines, (2) per-scene optimization based neural implicit reconstruction methods, (3) generalizable neural rendering methods, and (4) generalizable neural implicit reconstructions. 
Best scores are in \textbf{bold} and second best are \underline{underlined}.
} %
\label{table::fewviewchamfer}
\end{table*}

\customparagraph{Implementation details} 
We implement our model in PyTorch~\cite{imambi2021pytorch} and PyTorch Lightning~\cite{Falcon_PyTorch_Lightning_2019}.
During training, we use an image resolution of $640 \times 512$ and set the number of source images to $N = 4$. 
We train our model for 16 epochs using Adam~\cite{kingma2015adam} on one A100 GPU. %
The learning rate is set to $10^{-4}$. 
The ray number sampled per batch and the batch size are set to 1024 and 2, respectively. 
Similar to other volume rendering methods~\cite{mildenhall2021nerf,wang2021neus}, we use a hierarchical sampling strategy in both training and testing. 
We first uniformly sample $N_{\text{coarse}}$ points on the ray and then conduct importance sampling to sample another $N_{\text{fine}}$ points on top of the coarse probability estimation. 
We set $N_{\text{coarse}}=64$ and $N_{\text{fine}}=64$ during our experiments. 
For global feature volume $\mathbf{F}_v$, we set the resolution as $K=96$. 
During testing, we set the image resolution to $800\times600$.

\customparagraph{Baselines} 
We mainly compare our method with: (1)  SparseNeuS \cite{long2022sparseneus}, the state-of-the-art generalizable neural implicit reconstruction method; note that we report reproduced results using their official repository and the released model checkpoint; (2) generalizable neural rendering methods ~\cite{chen2021mvsnerf,yu2021pixelnerf,wang2021ibrnet}; (3) per-scene optimization based neural implicit reconstruction methods~\cite{yariv2020multiview,yariv2021volume,wang2021neus,oechsle2021unisurf}; (4) MVS methods~\cite{schonberger2016pixelwise,yao2018mvsnet}. We train MVSNet~\cite{yao2018mvsnet} with our training split for 16 epochs. Note that MVS methods are different from neural implicit reconstruction in that they do not implicitly model scene parameters, \eg, SDF, SRDF, and the state-of-the-art MVS methods are unable to render novel views. Similar to \cite{long2022sparseneus,wang2021neus}, we report them as a baseline. %

\subsection{Evaluation Results}
\label{sec:experiments}

\customparagraph{Sparse View Reconstruction on DTU} 
On DTU~\cite{aanaes2016_dtu}, we conduct sparse reconstruction with only 3 views. 
For a fair comparison, we adopt the same image sets and evaluation process as used in SparseNeuS~\cite{long2022sparseneus}. 
To calculate SRDF, we define a virtual rendering viewpoint corresponding to each view, which is generated by shifting the original camera coordinate frame for $d=25mm$ along its $x$-axis. 
After rendering the depth maps, we adopt TSDF fusion~\cite{curless1996volumetric} to fuse the depth maps in a volume with a voxel size of $1.5mm$, and then use Marching Cube~\cite{lorensen1987marching} to extract the mesh. 
As shown in Table~\ref{table::fewviewchamfer}, our method outperforms the state-of-the-art neural implicit reconstruction method SparseNeuS~\cite{long2022sparseneus} by 30\%. 
As for qualitative visualization shown in Fig.~\ref{fig:sparseview_mesh}, %
our method generates finer details and sharper boundaries than SparseNeuS. 
Compared with MVS methods~\cite{schonberger2016pixelwise,yao2018mvsnet}, we observe that our method outperforms the traditional MVS method COLMAP~\cite{schonberger2016pixelwise} by about 10\% but is a little worse than MVSNet~\cite{yao2018mvsnet}.%

\begin{figure*}[!ht]
    \centering
    \includegraphics[width=1.0\textwidth, page=4]{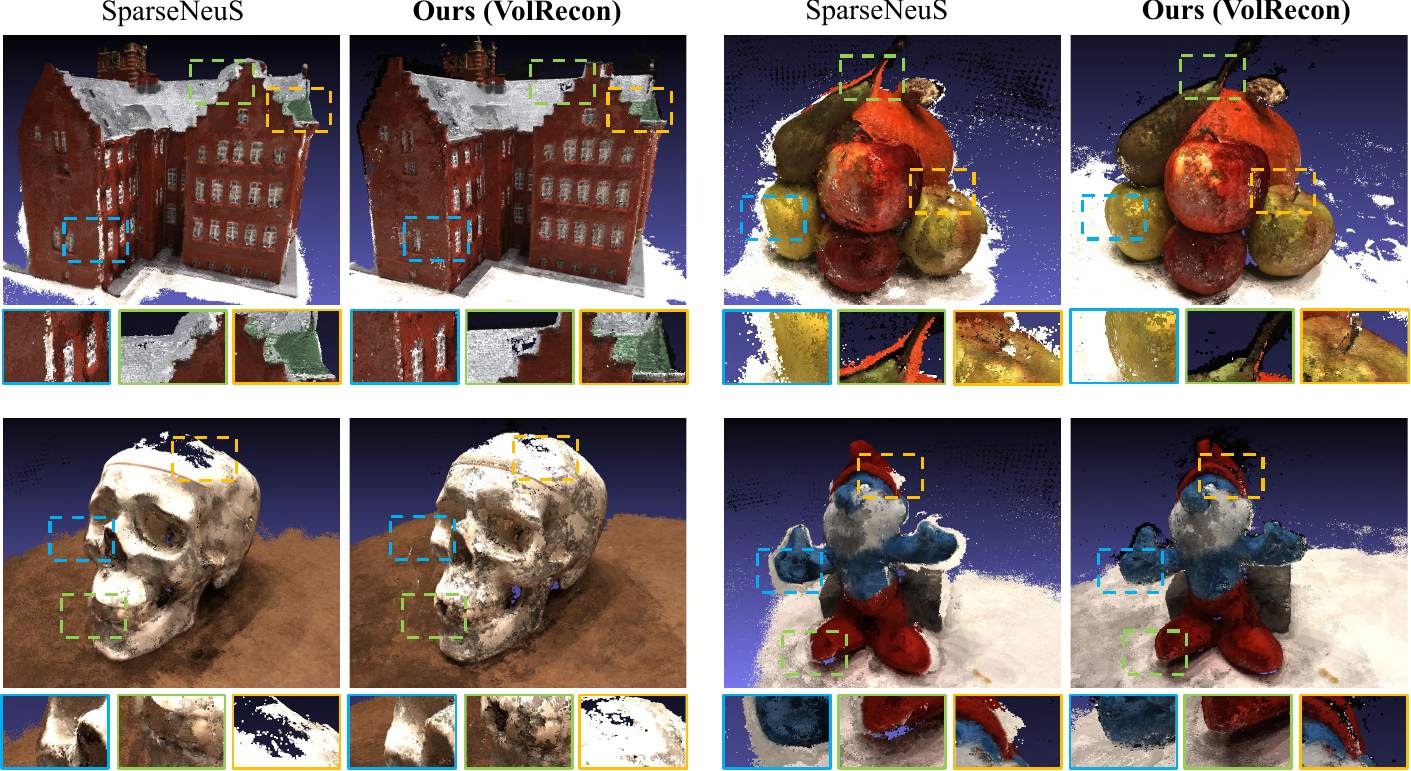}
    \caption{\textit{Point cloud} comparison of \textbf{full view} reconstruction on the DTU dataset~\cite{aanaes2016_dtu}. Compared with SparseNeuS~\cite{long2022sparseneus}, our method (VolRecon) reconstructs better point clouds,~\eg sharper boundary, as the steeple in the top left and pear stalk in the top right, and more complete representation,~\eg fewer holes, as skull head top in bottom left, foot in the bottom right. Note that each 3D point is projected with the rendered color and depth from each viewpoint. %
    Best viewed on a screen when zoomed in. 
    }
\label{fig:fullview_pcd_compare}
\end{figure*}

\customparagraph{Depth map evaluation on DTU} 
In this experiment, we compare depth estimation with SparseNeuS~\cite{long2022sparseneus} and MVSNet~\cite{yao2018mvsnet} by evaluating all views in each scan. %
For each reference view, 
we use 4 source views with the highest view selection scores according to \cite{yao2018mvsnet} for depth rendering. 
For SparseNeuS~\cite{long2022sparseneus}, we set the image resolution to $800\times600$ and render the depth similarly to our method. 
For MVSNet~\cite{yao2018mvsnet}, for a relatively fair comparison, we set the image resolution to $1600\times1184$ since the output depth is downsampled to $1/4$ resolution. 
As shown in Table~\ref{table::fullview_depthmetric}, our method achieves better performance in all the metrics than MVSNet and SparseNeuS. 

\begin{table}[th]
\resizebox{0.48\textwidth}{!}{%
\renewcommand{\arraystretch}{1.1}
\begin{tabular}{lrrrrr} \hline 
Method  & \multicolumn{1}{l}{$<1\uparrow$} & \multicolumn{1}{l}{$<2\uparrow$} & \multicolumn{1}{l}{$<4\uparrow$} & \multicolumn{1}{l}{Abs. $\downarrow$} & \multicolumn{1}{l}{Rel. $\downarrow$} \\ \hline
MVSNet \cite{yao2018mvsnet} & 29.95 & 52.82 & 72.33 & 13.62  & 1.67 \\ \hline
SparseNeuS \cite{long2022sparseneus} & 38.60 & 56.28  & 68.63  & 21.85 & 2.68 \\                                     
\textbf{Ours (VolRecon)} & \textbf{44.22}   & \textbf{65.62} & \textbf{80.19} & \textbf{7.87} & \textbf{1.00}  \\ \hline   
\end{tabular}
}
\caption{Depth map evaluation results on DTU \cite{aanaes2016_dtu}. The result of mean absolute error (Abs.) is in millimeters. The results of threshold percentage ($<1mm$, $<2mm$, $<4mm$) and mean absolute relative error (Rel.) are in percentage (\%). Best scores are in \textbf{bold}. }
\label{table::fullview_depthmetric}
\end{table}

\customparagraph{Full View Reconstruction on DTU}
Based on the depth maps of all the views, we further evaluate 3D reconstruction quality. For a fair comparison, we follow the MVS methods to fuse all 49 depth maps of each scan into one point cloud~\cite{galliani_2015_gipuma,yao2018mvsnet}. 
As shown in Table~\ref{table::fullview_point}, our method performs better than SparseNeuS and achieves comparable accuracy as MVSNet. 
As shown in Fig.~\ref{fig:fullview_pcd_compare}, compared with SparseNeuS, our method shows sharper boundary and fewer holes. %

\begin{table}[h]
\centering
\setlength{\belowcaptionskip}{-0.4cm}

\resizebox{0.38\textwidth}{!}{%
\renewcommand{\arraystretch}{1.1}
\begin{tabular}{lrrr} \hline 
Method  & \multicolumn{1}{l}{Acc.$\downarrow$} & \multicolumn{1}{l}{Comp.$\downarrow$} & \multicolumn{1}{l}{Chamfer$\downarrow$}  \\ \hline
MVSNet \cite{yao2018mvsnet} & \underline{0.55} & \textbf{0.59} & \textbf{0.57} \\ \hline
SparseNeuS \cite{long2022sparseneus} & 0.75 & 0.76  & 0.76  \\                                     
\textbf{Ours (VolRecon)} & \textbf{0.55} & \underline{0.66} & \underline{0.60}  \\ \hline   
\end{tabular}
}
\caption{Point cloud evaluation on DTU~\cite{aanaes2016_dtu}. For Accuracy (Acc.), Completeness (Comp.), and Chamfer distance, lower is better. Best scores are in \textbf{bold} and second best are \underline{underlined}. }
\label{table::fullview_point}
\end{table}

\begin{figure*}[!ht]
    \centering
    \setlength{\abovecaptionskip}{-0.1cm}
    \includegraphics[width=1.0\textwidth, page=5]{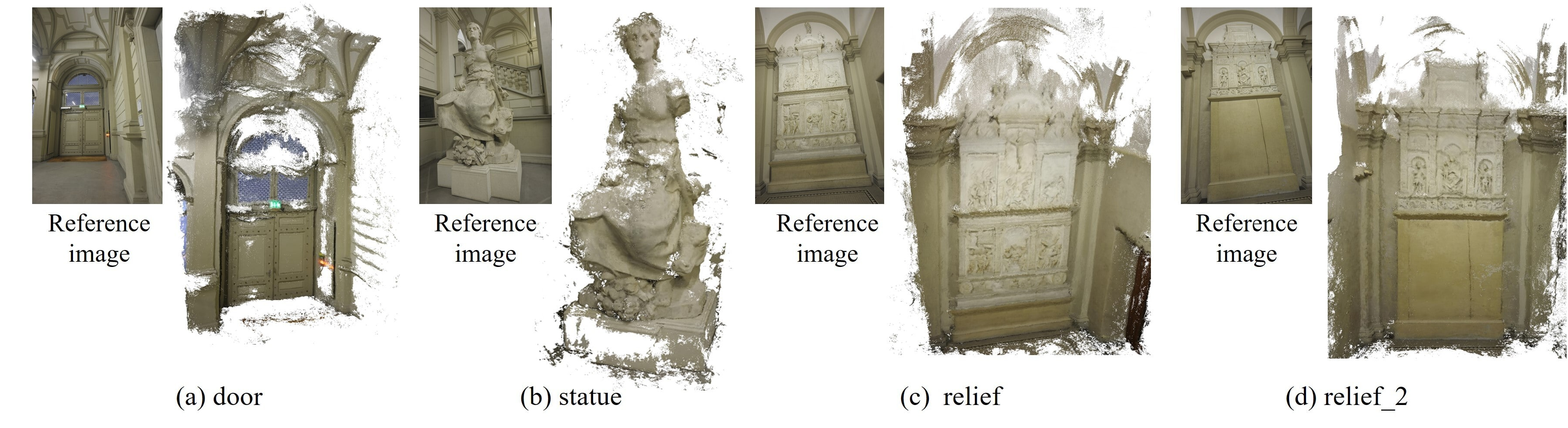}

    \caption{\textbf{Generalization ability} of VolRecon. Our model trained on DTU~\cite{aanaes2016_dtu} generalizes well the to large-scale strong viewpoint variation benchmark ETH3D~\cite{2017eth3d} without finetuning. Best viewed on a screen when zoomed in. %
    }

\label{fig:eval_eth3d}
\end{figure*}

\customparagraph{Generalization on ETH3D} 
To validate the generalization ability of our method, we directly test our model, pretrained using the DTU benchmark~\cite{aanaes2016_dtu}, on the ETH3D~\cite{2017eth3d} benchmark. 
We choose 4 scenes for testing: \textit{door}, \textit{statue}, \textit{relief}, and \textit{relief\_2}, which have 6, 11, 31, and 31 images, respectively.
Compared with DTU, the scale of the scenes increases about $10\times$. 
These large-scale scenes are not suitable to use TSDF fusion~\cite{curless1996volumetric} due to its limited voxel resolution. We render the depth maps and then fuse them into a point cloud~\cite{yao2018mvsnet} for each scene. 
As shown in Fig.~\ref{fig:eval_eth3d}, our method reconstructs large-scale scenes with high quality, which demonstrates that our method has good generalization capability.

\subsection{Ablation Study}
We conduct ablation studies to analyze the effectiveness of different components in our model.
All the experiments are done on the DTU benchmark~\cite{aanaes2016_dtu}. We summarize the results of the first three experiments on the sparse view ($N=3$) reconstruction, depth map evaluation, and full view reconstruction in Table~\ref{table::ablation_three}. 

\customparagraph{Ray Transformer}
By default, a ray transformer enables each point to attend to the features of other points on the ray. 
Then we remove the ray transformer and directly use the unattended features to predict SRDF. 
As shown in Table~\ref{table::ablation_three}, the performance drops in all the experiments. 
Without the ray transformer, the SRDF prediction only uses the local information of each point, which is not enough to accurately find the surface location along the ray.

\customparagraph{Global Feature Volume}
By default, we build a coarse global feature volume to encode global shape priors. We compare with not using global feature volume. 
The performance becomes worse. We conjecture that the local information from projection features is not enough to accurately locate the surface along a ray. The global feature volume provides global shape priors that are helpful for geometry estimation. 

\customparagraph{Depth Loss}
We remove the depth loss $\mathcal{L}_{\text {depth}}$ during training and observe that the reconstruction quality drops. 
Though, in sparse view reconstruction, our method still performs comparably to the SparseNeuS~\cite{long2022sparseneus} and is better than MVSNeRF~\cite{chen2021mvsnerf} and IBRNet~\cite{wang2021ibrnet}, as shown in Table~\ref{table::fewviewchamfer}. 
Many works find that only using pixel color loss $\mathcal{L}_{\text {color}}$ produces bad geometry in novel view synthesis~\cite{wei2021nerfingmvs}, especially in areas with little texture or repetitive patterns. 
Therefore, many implicit reconstruction methods use careful geometry initialization~\cite{yariv2021volume,wang2021neus} and geometric priors such as depth maps~\cite{yu2022monosdf}, normals~\cite{yu2022monosdf,wang2022neuris}, and sparse point clouds~\cite{fu2022geoneus} to provide more geometric supervision. 
Other methods~\cite{darmon2022improving,fu2022geoneus} use patch loss, which is common in %
unsupervised depth estimation methods~\cite{godard2017unsupervised} to provide more robust self-supervision in geometry than pixel color loss. 

\customparagraph{Number of Views}
We vary the number of views $N$ in sparse view reconstruction and summarize the results in Table~\ref{table::ablation_numofview}. 
The reconstruction quality gradually improves with more images. Multi-view information enlarges the observed areas and helps to alleviate problems such as occlusions. 

\begin{table*}[ht]
\centering

\resizebox{0.8\textwidth}{!}{%
\renewcommand{\arraystretch}{1.1}
\begin{tabular}{l|c|ccccc|c} \hline
\multirow{2}{*}{Method}         & Sparse View Recon. & \multicolumn{5}{c|}{Depth Map Eval.} & Full View Recon. \\ 
 \cline{2-8}
& Chamfer$\downarrow$ & $<1\uparrow$ & $<2\uparrow$ &  $<4\uparrow$ & Abs.$\downarrow$ & Rel. $\downarrow$ & Chamfer$\downarrow$ \\
\hline
w/o Ray Trans. & 1.79 & 39.20   & 60.73 & 77.38 & 8.80 & 1.12 & 0.66 \\
w/o $\mathbf{F}_v$ & 1.83 & 23.29   & 40.67 & 59.64 & 14.90 & 1.92 & 0.78 \\
w/o $\mathcal{L}_{\text {depth}}$ & 2.04 & 12.84   & 22.55 & 34.91 & 35.00 & 4.41 & 1.24 \\ 
\textbf{\textbf{Ours (VolRecon)}} & \textbf{1.38} & \textbf{44.22}   & \textbf{65.62} & \textbf{80.19} & \textbf{7.87} & \textbf{1.00} & \textbf{0.60}  \\ \hline  
\end{tabular}
}
\caption{Ablation study of ray transformer, global feature volume, and depth loss on DTU \cite{aanaes2016_dtu} dataset. Best scores are in \textbf{bold}.
}
\label{table::ablation_three}
\end{table*}

\begin{table}[h]
\centering
\begin{widetable}{0.6\columnwidth}{cc} \hline 
Number of Views  & Chamfer $\downarrow$ \\ \hline         
2 & 1.72  \\
3 & 1.38 \\
4 & 1.35 \\
5 & \textbf{1.33} \\
\hline   
\end{widetable}
\caption{Ablation study of number of views on DTU benchmark~\cite{aanaes2016_dtu} dataset. The Chamfer distance is reported (lower is better). Best score is in \textbf{bold}. 
}
\label{table::ablation_numofview}
\end{table}

\section{Limitations \& Future Work}

There are two limitations of our method. First, the rendering efficiency of our method is limited, which is a common problem in other volume rendering-based methods~\cite{mildenhall2021nerf,chen2021mvsnerf,wang2021ibrnet}. It takes about $30$s to render an image and depth map with a resolution of $800\times600$. %
Second, our current model is not suitable for reconstructing very large-scale scenes. 
The low resolution of our global feature volume results in a decrease in representation performance when the scale of the scene increases. 
While increasing the resolution of the global feature volume is a potential solution, this will increase memory consumption.
Instead, we believe it will be a promising direction to reconstruct progressively in small local volumes like NeuralRecon~\cite{sun2021neuralrecon}. 
To implement this strategy, given a rendering viewpoint, we will select several source views~\cite{schonberger2016pixelwise,yao2018mvsnet,duzceker2021deepvideomvs} to build a local bounding volume that encloses their view frustums. This will effectively limit the space to a reasonable size and allow us to apply our method within the local region.

\section{Conclusion}
\label{sec:conclusion}

We introduced VolRecon, a novel generalizable implicit reconstruction method with SRDF. 
Our method incorporates a view transformer for aggregating multi-view features and a ray transformer for computing SRDF values of all the points along a ray to find the surface location. 
By utilizing both projection features and volume features, our approach is able to combine local information and global shape prior, and thus produce reconstructions with fine details and of high quality. 
Our method outperforms the state-of-the-art generalizable neural implicit reconstruction methods on DTU by a large margin. Furthermore, experiments on ETH3D without any fine-tuning demonstrate good generalization ability on large-scale scenes. 

\textbf{Acknowledgements} This work was supported in part by the Swiss National Science Foundation via the Sinergia grant CRSII5-180359.

\clearpage
\noindent{\fontsize{14}{14} \textbf{Appendix}}

\section{Volume Rendering of SRDF and SDF}
\begin{figure}[h]
\setlength{\belowcaptionskip}{-0.6cm}
\centering
{\includegraphics[width=1.0\linewidth, page=6]{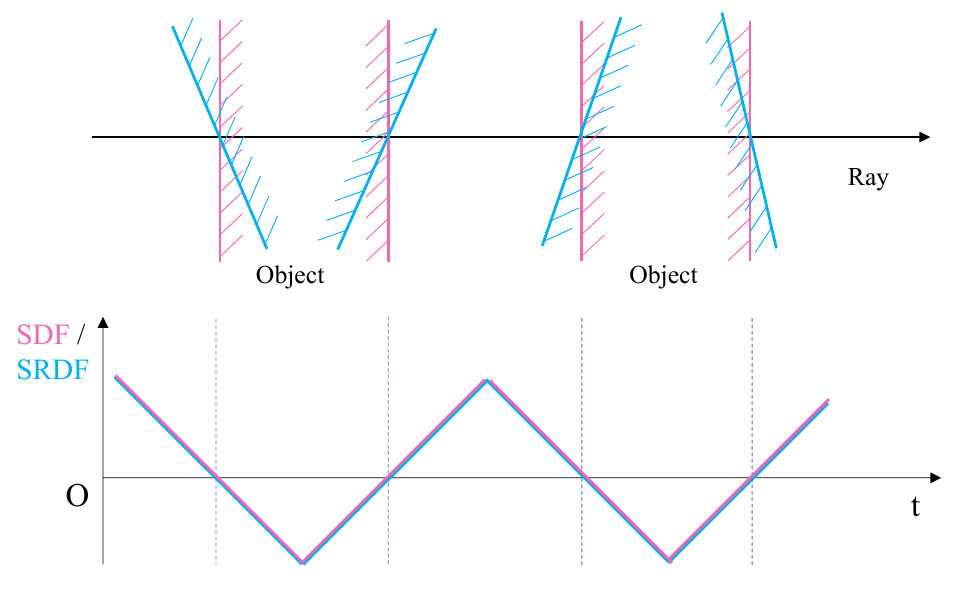}}
\caption{
A horizontal ray penetrating surfaces (shadowed lines), in the case of vertical (\textcolor{pink}{pink}) and random (\textcolor{blue}{blue}) angle, top.
SRDF (\textcolor{blue}{blue}) is irrelevant to the incidence angle and is equal to the SDF where the surface is vertical to the ray (\textcolor{pink}{pink}), bottom.
}\label{fig::srdf_sdf}
\end{figure}

We volume render the Signed Ray Distance Functions (SRDF) based on the volume rendering theory for Signed Distance Functions (SDF) from NeuS~\cite{wang2021neus}. 
As shown in Fig.~\ref{fig::srdf_sdf}, we consider multiple surface intersections (shadowed lines) along the ray with several intersection points. 
For SRDF, we set the surface normal vectors to be random at the intersection points (\textcolor{pink}{pink}) since the distribution of SRDF is irrelevant to the surface normal. 
For SDF, we set the surfaces to be perpendicular (\textcolor{blue}{blue}) to the ray at the intersection points. 
In this case, the distributions of SRDF and SDF values are the same. 
Therefore, the distribution of SRDF along the ray is the same as that of SDF with the surfaces perpendicular to the ray direction at the same surface intersection points. 
Thus we can adopt the way of volume rendering SDF~\cite{wang2021neus} to volume render SRDF. 

\section{Global Feature Volume}
\begin{figure}[ht]
\setlength{\belowcaptionskip}{-0.375cm}

    \centering
    \includegraphics[width=1.0\linewidth]{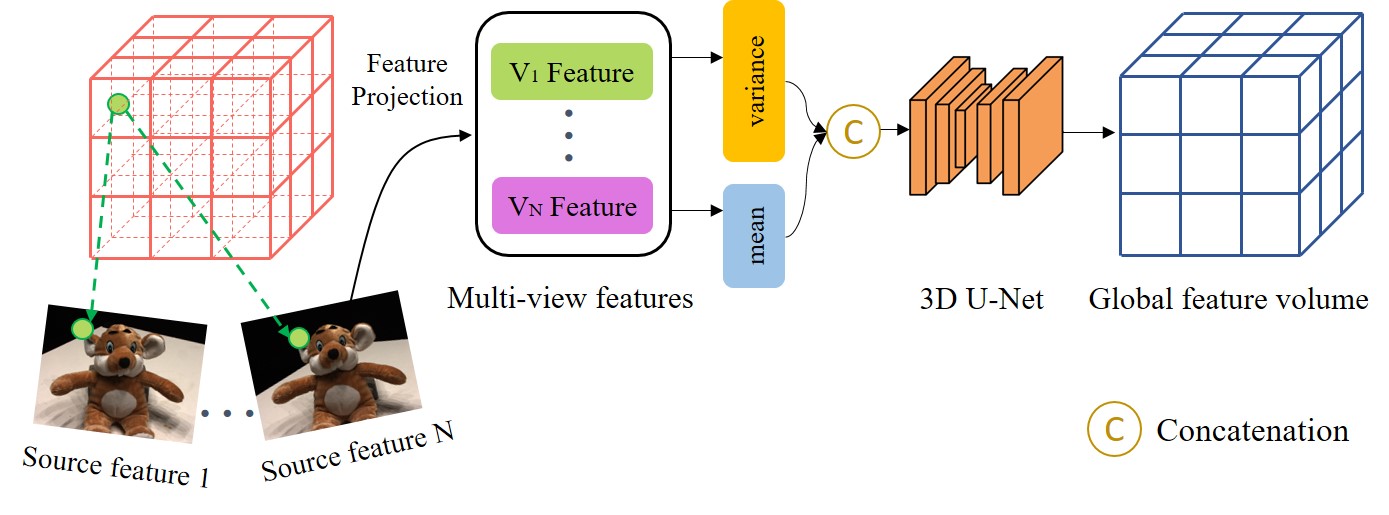}
    
    \caption{Pipeline of constructing global feature volume. Best viewed on a screen when zoomed in. } 

\label{fig:golbal_feature_volume}
\end{figure}

Recall that we construct a global feature volume $\mathbf{F}_v$ to get global information. 
After dividing the bounding volume of the scene into $K^3$ voxels, we project the center point of each voxel onto the feature map of each source view and obtain the feature with bilinear interpolation. 
Then we compute the mean and variance of $N$ features for each voxel and concatenate them as the voxel feature. 
Finally, we use a 3D U-Net\cite{ronneberger2015u} for regularization and get the global feature volume $\mathbf{F}_v$. 
The pipeline is shown in Fig.~\ref{fig:golbal_feature_volume}. 

\section{Point Cloud Reconstruction}
Recall that we reconstruct the scene with both TSDF fusion~\cite{curless1996volumetric} and point cloud fusion. 
For point cloud reconstruction, we follow the MVS method~\cite{yao2018mvsnet}. 
Before fusing the depth maps, we filter out unreliable depth estimates with geometric consistency filtering, which measures the consistency of depth estimates among multiple views. 
For each pixel $\mathbf{p}$ in the reference view, we project it with its depth $d_0$, to a pixel $\mathbf{p}_i$ in the $i$-th source view. 
After retrieving its depth $d_i$ in the source view with interpolation, we re-project $\mathbf{p}_i$ back into the reference view, and retrieve the depth $d_\textrm{reproj}$ at this location,  $\mathbf{p}_\textrm{reproj}$. 
We consider pixel $\mathbf{p}$ and its depth $d_0$ as consistent to the $i$-th source view, 
if the distances, in image space and depth, between the original estimate and its re-projection satisfy: 
\begin{equation}
    |\mathbf{p}_\textrm{reproj}-\mathbf{p}| < \delta, |d_\textrm{reproj}-d_0|/d_0 < \varepsilon,
\end{equation}
where $\delta$ and $\varepsilon$ are two thresholds. We set $\delta=1$ and $\varepsilon=0.01$, which are the default parameters from MVSNet~\cite{yao2018mvsnet}. %
Finally, we accept the estimations as reliable if they are consistent in at least $N_\textrm{geo}$ source views.

\section{Generalization on ETH3D}
\begin{figure*}[!ht]
    \centering
    \includegraphics[width=1.0\textwidth, page=1]{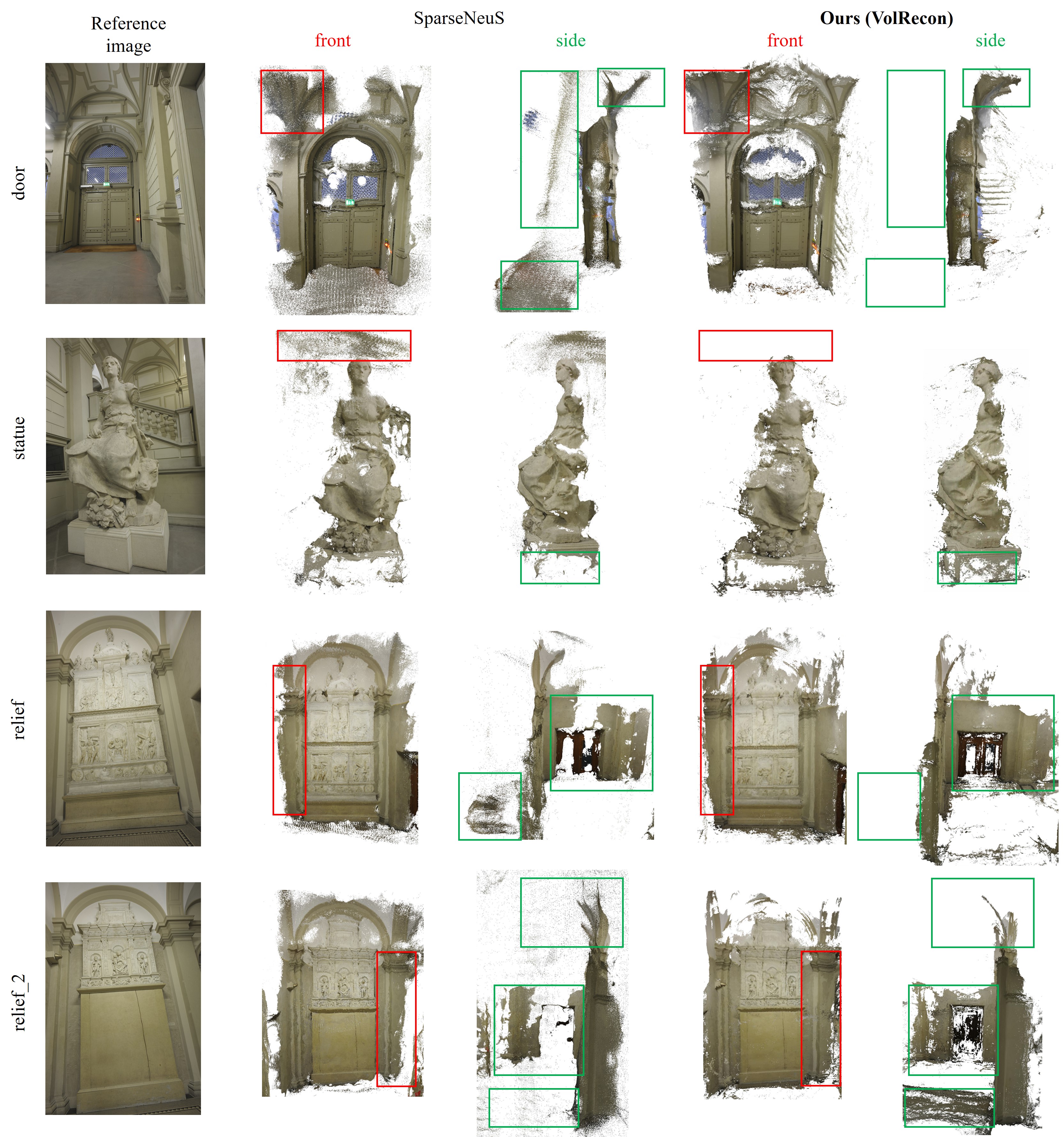}
    \caption{Point cloud visualization of reconstructions on ETH3D~\cite{2017eth3d} benchmark. %
    Compared with SparseNeuS~\cite{long2022sparseneus}, our method produces better reconstruction with less noise (\eg, ground of scene \textit{door} and top of scene \textit{statue}) and higher completeness (fewer holes, \eg, wall of scene \textit{relief} and \textit{relief\_2}). 
    Best viewed on a screen when zoomed in. 
    }
\label{fig:eth3d_compare}
\end{figure*}

We compare the generalization ability of SparseNeuS~\cite{long2022sparseneus} and our method on the ETH3D~\cite{2017eth3d} benchmark. %
Recall that we directly test our model, pretrained on DTU~\cite{aanaes2016_dtu}, on ETH3D. %
For a fair comparison, we test the DTU pre-trained SparseNeuS~\cite{long2022sparseneus} with the same dataset settings and point cloud reconstruction process. 
As shown in Fig.~\ref{fig:eth3d_compare}, our method reconstructs the scenes with less noise and higher completeness (fewer holes) than SparseNeuS~\cite{long2022sparseneus}. 
This further demonstrates that our method has good generalization capability for large-scale scenes.

\section{Baselines with Depth Supervision}
Due to the ambiguity between appearance and geometry in NeRF~\cite{zhang2020nerf}, recent methods~\cite{fu2022geoneus,wei2021nerfingmvs,zhang2022critical} mainly add additional 3D supervision, \eg depth and normal, into baselines (VolSDF, NeuS) to compare with naive baselines (pixel color loss only). 

\begin{figure}[!h]
\setlength{\belowcaptionskip}{-0.375cm}
    \centering
    \includegraphics[width=0.95\columnwidth,  page=7]{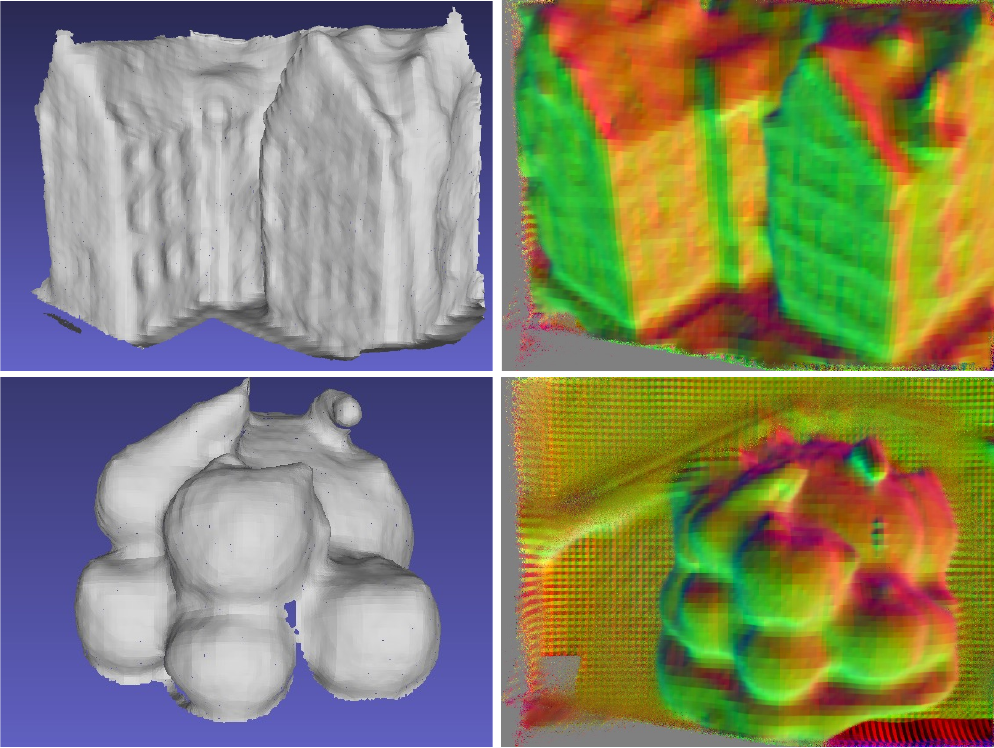}
    
    \caption{Visualization of reconstructed mesh and rendered normal map for $\text{SparseNeuS}_d$. Best viewed when zoomed in.} 

\label{fig:sparseneusdepth}
\end{figure}

For a fair comparison, we trained a SparseNeuS model while adding the depth loss (denoted $\text{SparseNeuS}_d$) with default settings and the same loss coefficient as ours. 
Besides, we remove depth loss in VolRecon and denote it as VolRecon*. 
VolRecon* performs slightly worse with SparseNeuS* (2.04\footnote{Chamfer distance, the lower the better, same below} v.s. 1.96) in sparse view reconstruction. We conjecture the reason to be we not using a shape initialization as SparseNeuS~\cite{wang2021neus,long2022sparseneus}. However, $\text{SparseNeuS}_d$ still reconstructs over-smoothed surfaces and even performs worse (4.22), Fig.~\ref{fig:sparseneusdepth}, while ours performs better with depth supervision. Furthermore, we noticed that the grid-like pattern persists in the rendered normal map due to limited voxel resolution. 

\section{Novel View Synthesis of VolRecon}
\begin{figure*}[!h]
\setlength{\belowcaptionskip}{-0.375cm}
    \centering
    \includegraphics[width=0.99\textwidth]{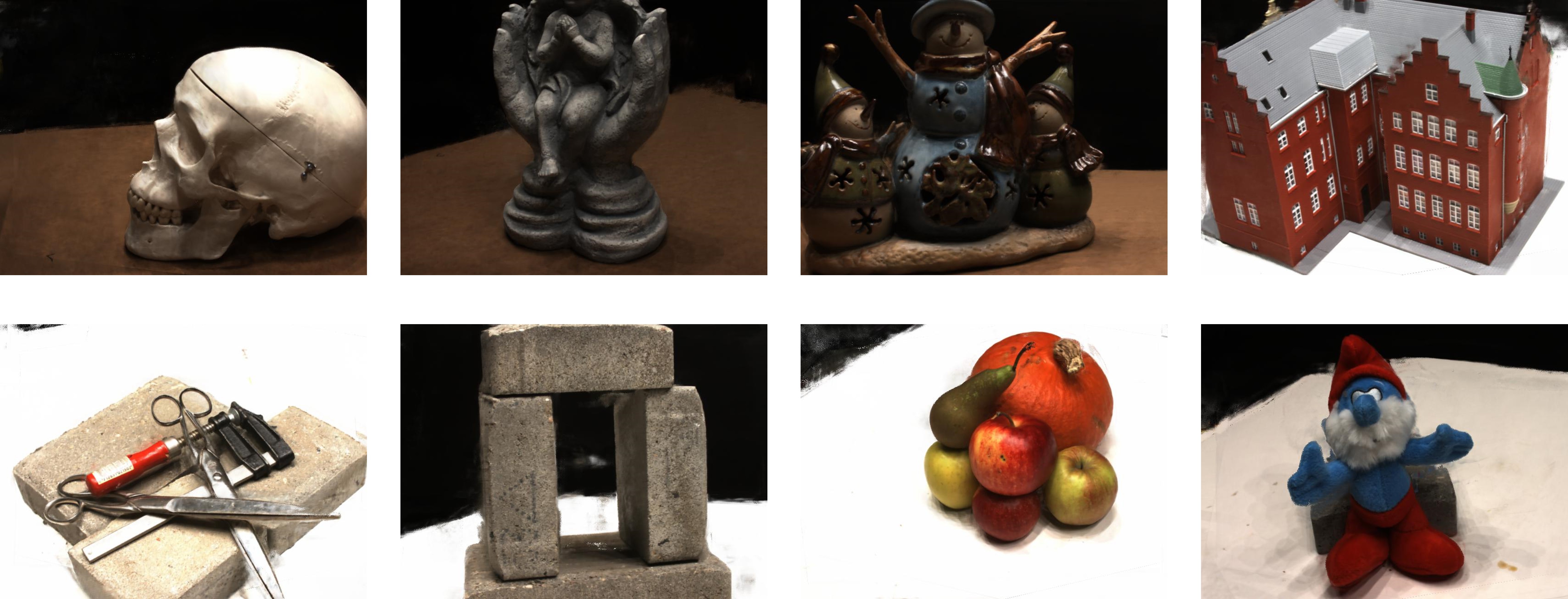}
    
    \caption{Visualization of novel view synthesis. Best viewed when zoomed in. } 

\label{fig:render_rgb}
\end{figure*}

We report novel view synthesis results on DTU dataset~\cite{aanaes2016_dtu} in Table~\ref{table::novel_view_synthesis}, where we use the same dataset setting as full view reconstruction and render each view with 4 source views only. Qualitative results are shown in Fig.~\ref{fig:render_rgb}. 

\begin{table}[h]
\centering

\begin{widetable}{0.6\columnwidth}{cccc} \hline 
 Method & PSNR $\uparrow$  & MSE $\downarrow$  & SSIM $\uparrow$ \\ \hline         
Ours & 15.37 & 0.04 & 0.56  \\
\hline   
\end{widetable}
\caption{Quantitative results of novel view synthesis on DTU~\cite{aanaes2016_dtu}. Each view is rendered with 4 source views only. 
}
\label{table::novel_view_synthesis}
\end{table}

\section{Point Cloud Visualization on DTU}
\begin{figure*}[!ht]
    \centering
    \includegraphics[width=1.0\textwidth, page=8]{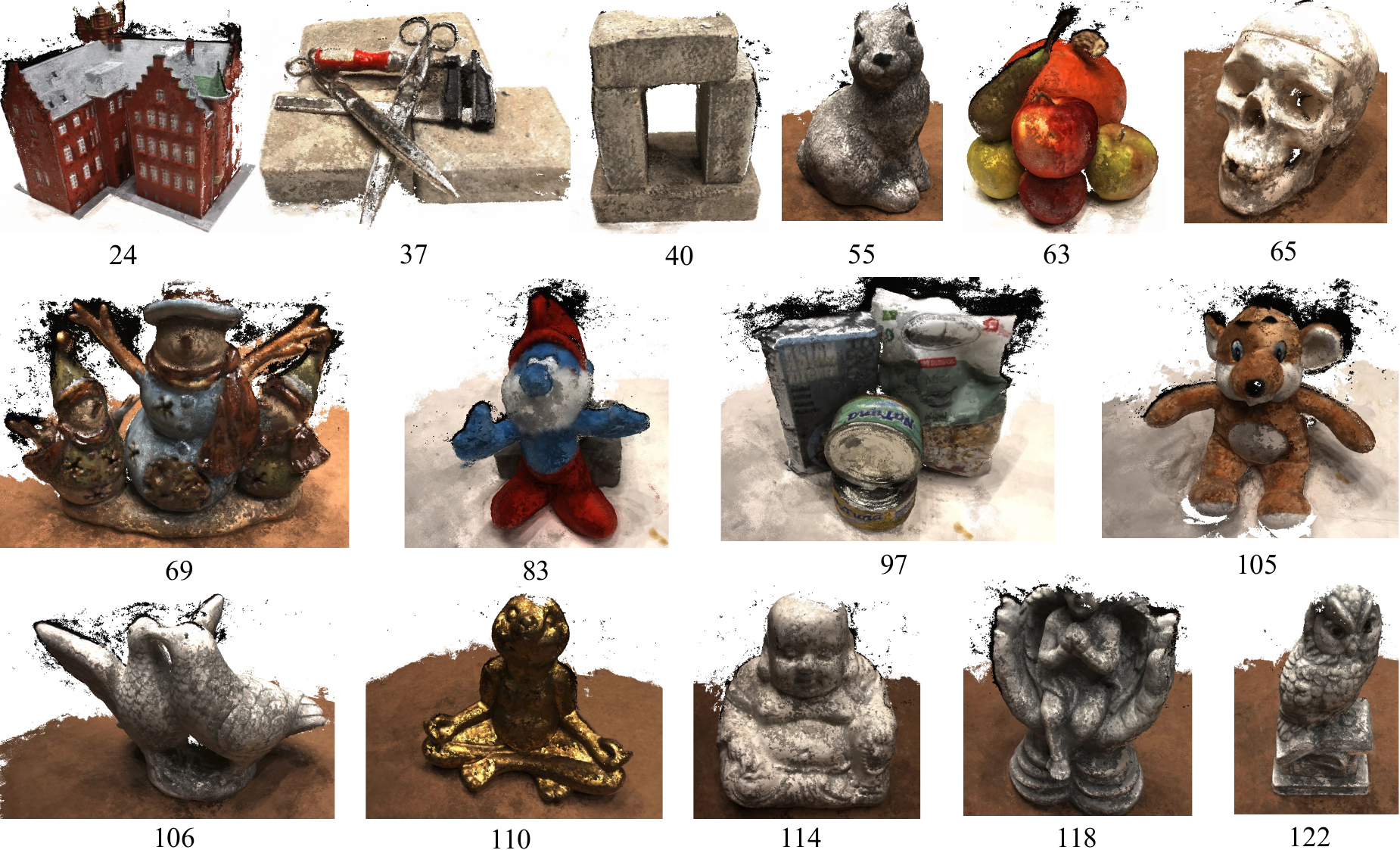}
    \caption{Point cloud visualization of the full view reconstructions on the DTU dataset~\cite{aanaes2016_dtu}. Best viewed on a screen when zoomed in. 
    }
\label{fig:fullview_pcd}
\end{figure*}

We visualize all the reconstructed point clouds of the full view reconstructions on the DTU dataset~\cite{aanaes2016_dtu} in Fig.~\ref{fig:fullview_pcd}. 

{\small
\bibliographystyle{ieee_fullname}
\bibliography{11_references}
}

\end{document}